\documentclass{article}

    \usepackage[preprint]{neurips_2021}

\pdfoutput=1
\usepackage[utf8]{inputenc} 
\usepackage[T1]{fontenc}    
\usepackage{url}            
\usepackage{booktabs}       
\usepackage{amsfonts}       
\usepackage{nicefrac}       
\usepackage{microtype}      
\usepackage{xcolor}         

\usepackage[pdftex]{graphicx}
\usepackage{wrapfig}
\usepackage{amsmath} 
\usepackage[ruled,vlined]{algorithm2e}
\usepackage[hidelinks]{hyperref}

\usepackage{graphicx}
\usepackage{subfigure}
\usepackage{booktabs,makecell,multirow}

\title{A new semi-supervised inductive transfer learning framework: Co-Transfer}

\author{%
  Zhe Yuan,$^{2}$ \qquad Yimin Wen$^{1, 2}$ 
  \\
\begin{tabular}[t]{c@{\extracolsep{3em}}c}
 \\ $^{1}$Guangxi Key Laboratory of Image and Graphic intelligent processing, Guilin University of \\ Electronic Technology, Guilin 541004, China \\ $^{2}$School of Computer Science and Information Safety, Guilin University of \\ Electronic Technology, Guilin 541004, China \\
\texttt{zyuan16888@gmail.com} \quad \texttt{ymwen@guet.edu.cn}
\end{tabular}
}

\begin{document}

\maketitle

\begin{abstract}
In many practical data mining scenarios, such as network intrusion detection, Twitter spam detection, and computer-aided diagnosis, a source domain that is different from but related to a target domain is very common. In addition, a large amount of unlabeled data is available in both source and target domains, but labeling each of them is difficult, expensive, time-consuming, and sometime unnecessary. Therefore, it is very important and worthwhile to fully explore the labeled and unlabeled data in source and target domains to settle the task in target domain. In this paper, a new semi-supervised inductive transfer learning framework, named \emph{Co-Transfer} is proposed. Co-Transfer first generates three TrAdaBoost classifiers for transfer learning from the source domain to the target domain, and meanwhile another three TrAdaBoost classifiers are generated for transfer learning from the target domain to the source domain, using bootstraped samples from the original labeled data. In each round of co-transfer, each group of TrAdaBoost classifiers are refined using the carefully labeled data. Finally, the group of TrAdaBoost classifiers learned to transfer from the source domain to the target domain produce the final hypothesis. Experiments results illustrate Co-Transfer can effectively exploit and reuse the labeled and unlabeled data in source and target domains.
\end{abstract}

\section{Introduction}
Transfer learning has been being widely utilized to transfer knowledge from a source domain to a related target domain even they are in different feature spaces or with different distributions \citep{1, 2}. According to Yang \citep{1}, transfer learning can be categorized under three sub-settings, inductive transfer learning, transductive transfer learning, and unsupervised transfer learning. In the inductive transfer learning setting, the data in target domain are all labeled no matter the data in source domain are labeled or not. In transductive transfer learning setting, the data in target domain are unlabeled while the data in source domain are labeled. While in unsupervised transfer learning setting, the data in target and source domains are all unlabeled. 

With different setting, semi-supervised transfer learning \citep{3, 4, 5, 6} has been concerned recently. It is utilized to handle many real applications in which only part of the data in target domain are labeled while the data in source domain are labeled or a model is pretrained in source domain. However, it is more common in many real application scenarios that a small amount of labeled data in source and target domains are available while a large amount of unlabeled data in source and target domains are also available. For example, in application of computer-aided diagnosis (CAD) systems \citep{7}, only a small amount of images can be carefully diagnosed by medical experts while a large number of unlabeled medical images are very easy to obtain. Due to aging or upgrading of equipments, the pattern of the medical images collected before may be different with the collected at the current time \citep{8, 9}. Other examples include network intrusion detection and Twitter spam detection. That is to say, the distributions of the data collected in two different time interval are different. Thus, the following challenge is how to transfer knowledge from the labeled and unlabeled data in source domain to help learning on the labeled and unlabeled data in target domain? 

In this paper, to handle the problem mentioned above, a new semi-supervised inductive transfer learning paradigm named Co-Transfer is proposed. It combines semi-supervised learning with instance-based transfer learning. Co-Transfer first generates three TrAdaBoost \citep{10} classifiers for transferring knowledge from the source domain to the target domain, and meanwhile another three TrAdaBoost classifiers are generated for transferring knowledge from the target domain to the source domain, using bootstraped samples from the original labeled data. In each round of co-transfer, each group of TrAdaBoost classifiers are refined using the newly labeled data, in which one part is labeled by itself and the other part is labeled by another group of TrAdaBoost classifiers. It is should be noted that the newly labeled samples are carefully selected under certain conditions, which is used in tri-training \citep{11}. Finally, the group of TrAdaBoost classifiers learned to transfer from the source domain to the target domain produce the final hypothesis via majority voting. Experiments on UCI datasets \citep{12} and text classification tasks verify Co-Transfer can effectively exploit and reuse the labeled and unlabeled data of the source and target domains.

The rest of this paper is organized as follows. Section 2 briefly reviews semi-supervised learning, instance-based transfer learning and semi-supervised transfer learning. Section 3 presents Co-Transfer. Section 4 reports the experimental results on UCI data sets and the text classification task. Finally, Section 5 concludes this paper.

\section{Related work}
In this section, we briefly review the related work of semi-supervised learning, instance-based transfer learning, and semi-supervised transfer learning.

\subsection{Semi-supervised learning}
For semi-supervised learning \citep{13, 14, semi_text_classification}, the main idea is to utilize a small number of labeled samples and a large amount of unlabeled samples to improve the performance of the learned hypothesis. One of the assumptions for semi-supervised learning is that the unlabeled examples hold the same distribution as that held by the labeled ones. There are many classical approaches have been proposed and can be mainly divided into several categories, like generative models \citep{15, 16}, wrapper methods \citep{17, 11, 18}, low-density separation models \citep{19}, and graph-based methods \citep{20} etc. Among them, the wrapper method is a type of simple and popular method, which first trains classifiers on the original labeled data, and then to utilize the predictions of the resulting classifiers to generate newly labeled data. The classifiers is then re-trained on these pseudo-labeled data in addition to the existing labeled data. 

Avrim Blum and Tom Mitchell proposed the Co-training algorithm \citep{17}, which requires two sufficient and redundant views. Co-training trains two classifiers separately on two different views, and uses the predictions of each classifier on unlabeled examples to augment the training set of the other. Dasgupta et al. \citep{21} have shown that when the requirement is met, the co-trained classifiers could make fewer generalization errors by maximizing their agreement over the unlabeled data. Since two sufficient and redundant views can hardly be met in most scenarios. Zhou et al. \citep{11} proposed the tri-training algorithm, which does not require two sufficient and redundant views. Later, Li et al. \citep{18} proposed the co-forest algorithm by extending tri-training to more classifiers. In tri-training and co-forest, to ensure that the newly labeled samples can bring positive effects, fuzzy set theory \citep{22} is introduced. Some other wrapper methods can be found in  \citep{14}.

\subsection{Instance-based transfer learning}
The intuitive idea of the instance-based inductive transfer methods is to reuse certain parts of the data in source domain together with the target domain data to train a high-accurate classifier. 

Dai et al. \citep{10} proposed TrAdaBoost, which is an extension of AdaBoost. TrAdaBoost attempts to iteratively re-weight the source domain data to reduce the effect of the \emph{bad} source data while encourage the \emph{good} source data to contribute more for target domain. In addition, they also theoretically analyzed that TrAdaBoost can converge well. Similarly, Kamishima et al. \citep{23} proposed a TrBagg method, which is an extension of bagging. The TrBagg training process consists of two stages: learning and filtering. In the learning stage, weak classifiers are trained on bootstraped samples from the target and source data sets. In the filtering phase, these weak classifiers are evaluated on the target domain data. Weak classifiers with low accuracy will be discarded while the weak classifiers with high accuracy will be selected to produce the final hypothesis. Later, Shi et al. \citep{24} proposed the COITL algorithm for inductive transfer learning by extending the co-training method for semi-supervised learning. The key idea of COITL is to replace the step of labeling the unlabeled examples by re-weighting the source domain data. In COITL, two base learners are trained on the target domain data, and each learner is refined using the weighted source domain examples predicted by the other. In addition, a number of other instance-based inductive transfer methods have been proposed to extend single source domain to multiple source domains \citep{25, 26, 27, 28}.

\subsection{Semi-supervised transfer learning}

Semi-supervised transfer learning has not received widespread attention, to the best of our knowledge, and only several related papers are found. According to these papers, semi-supervised transfer learning always aims to tackle learning on the labeled and unlabeled data in target domain. Due to different forms of using source domain, it is mainly divided into two categories: the source domain data is directly available and the source domain data is not available but a pre-trained models is available. 

Liu et al. \citep{4} proposed a tri-training based transfer learning algorithm (TriTransfer). In TriTransfer, three initial classifiers are generated from the labeled data in source domain and the originally labeled data in target domain, and then an unlabeled example in target domain is labeled and added to the labeled data for a classifier if other two classifiers agree on its label. After an expanded labeled data set is obtained, the classifier is re-trained. The final classifier is a weighted combination of the three classifiers. 

Tang et al. \citep{5} proposed a semi-supervised transfer learning algorithm to use Convolutional Neural Network (CNN) for Chinese character recognition. First, a CNN model is trained by massive labeled samples in source domain. Then the CNN model is fine-tuned by limited labeled samples in target domain. Finally, the fine-tuned CNN model is continuously trained with the unlabeled samples in target domain and kept fine-tuning the model with the labeled samples in target domain to minimize the loss. 

Wei et al. \citep{6} proposed a semi-supervised transfer learning algorithm for image rain removal. The network is therefore trained to adapt real unsupervised diverse rain types through transferring from the supervised synthesized rain by elaborately formulating the residual between an unlabeled rainy image and its expected network output as a specific parameterized rain streaks distribution. 

Abuduweili et al. \citep{3} proposed a semi-supervised transfer learning algorithm to utilize both the powerful pre-trained models from source domain as well as the labeled/unlabeled data in target domain. Adaptive knowledge consistency (AKC) on the examples between the source and target model is utilized to transfer knowledge from the pre-trained model and help generalize the target model, while adaptive representation consistency (ARC) on the target model between labeled and unlabeled examples is utilized to adjust the representation produced by the supervised learning to the real target domain.

\section{Co-Transfer}

\subsection{Problem formulation}

To formalize the definition of semi-supervised inductive transfer learning proposed in this paper, some notations are introduced.

In this paper, it is assumed that the source domain $D_S$ and the target domain $D_T$ have the same feature space but different distributions. $\mathcal{X}$ denotes the feature space, and $\mathcal{Y}=\{0,1\}$ denotes the label space. $x$ is a data instance. Both  $D_S$ and $D_T$ are divided into two parts: labeled part and unlabeled part. The labeled part includes a few labeled instances, $D_{SL}=\{(x_{1}^{S},y_{1}^{S}),\dots,(x_{l}^{S},y_{l}^{S})\}$ and $D_{TL}=\{(x_{1}^{T},y_{1}^{T}),\dots,(x_{p}^{T},y_{p}^{T})\}$, where $x_{i}^{S} \in \mathcal{X}$ is an instance in the source domain, and $y_{i}^{S} \in \mathcal{Y}$ is the corresponding class label, while $x_{i}^{T} \in \mathcal{X}$ is an instance in the target domain, and $y_{i}^{T} \in \mathcal{Y}$ is the corresponding class label. The unlabeled part includes unlabeled data, $D_{SU}=\{(x_{l+1}^{S},\dots,x_{l+m}^{S})\}$ and $D_{TU}=\{(x_{p+1}^{T},\dots,x_{p+n}^{T})\}$. Furthermore, a test data set $D_{Test}=\{(x_{1}^{T},\dots,x_{r}^{T})\}$, which has the same distribution as the target domain $D_T$, is available. For simplicity, in this paper, we set $l$=$p$. In addition, a large amount of unlabeled data in the source and target domain are available, usually $m \gg l$ and $n \gg p$. The objective of the proposed semi-supervised inductive transfer learning is to utilize $L=[D_{SL},D_{TL}]$ and $U=[D_{SU},D_{TU}]$ to learn a function $f:\mathcal{X}\rightarrow \mathcal{Y}$ on the target domain, such that $f$ can correctly predict the label of instances in $D_{Test}$.

\subsection{Co-Transfer}

In Co-Transfer, two ensemble classifiers $H^{0}=\{h_{1}^{0}$, $h_{2}^{0}$, $h_{3}^{0}\}$ and $H^{1}=\{h_{1}^{1}$, $h_{2}^{1}$, $h_{3}^{1}\}$ are initially trained. By TrAdaBoost \citep{10}, each base classifier $h_{i}^{0} (i=1,2,3)$ of $H^0$ is trained on the bootstraped samples from the original labeled data sets $L[1]$ and $L[0]$ for transfer learning from the target domain to the source domain. Correspondingly, each base classifier $h_{i}^{1} (i=1,2,3)$ of $H^1$ is trained on the bootstraped samples from the original labeled data sets $L[0]$ and $L[1]$ for transfer learning from the source domain to the target domain. Thus, the bootstrap strategy keeps the diversity between the base classifiers in  $H^0$ and $H^1$. The mutual transfer learning between the source and target domains like this will iterate many rounds. Each base classifier of $H^0$ and $H^1$ is then refined with the newly-labeled samples. 

In detail, in each iteration round of Co-Transfer, the unlabeled samples in $U[d]$ ($d=\{0,1\}$, $d=0$ \emph{corresponds to the source domain, while}  $d=1$ \emph{corresponds to the target domain}) are labeled and added to the original labeled data $L[d]$ as target data to retrain the classifier $h_{i}^{d}$ ($i$=$1,2,3$). An unlabeled sample in $U[d]$ can be labeled if the other two classifiers $H_{i}^{d}$=$h_{j}^{d}\cup h_{k}^{d}$ ($j\neq k \neq i$) agree on labeling this sample. The newly-labeled data set is named as $L_{i}^{d}$. Meanwhile, the unlabeled samples in $U[(d+1)\%2]$ are labeled and added to the original labeled data $L[(d+1)\%2]$ as source data to retrain $h_{i}^{d}$. An unlabeled sample in $U[(d+1)\%2]$ can be labeled if all the base classifiers in the ensemble $H^{(d+1)\%2}$ agree on labeling this sample. The newly-labeled data set is named as $L^{(d+1)\%2}$. In summary, $L^{(d+1)\%2} \cup L[(d+1)\%2]$ and $L_{i}^{d} \cup L[d]$ are employed as the source and target data respectively to retrain $h_{i}^{d}$ by TrAdaBoost. Note that all the unlabeled samples that are selected by $H^d$ or $H_{i}^{d}$ ($i=1,2,3$) are not removed from $U[d]$. Therefore, they might be selected again  in the following iterations. When none of $H^{d}$ ($d=0,1$) changes the iteration ends. The final hypothesis for the target domain is produced via majority voting by $H^1$.

Like Tri-training \citep{11}, the employment of the ensemble of classifiers $H^d$ and $H_{i}^{d}$ ($i=1,2,3$) not only serves as a simple way to avoid utilizing complicated confidence and transferable estimation methods but also makes the labeling of unlabeled data more reliable than a single classifier does. However, although the ensemble $H^d$ or $H_{i}^{d}$ ($i=1,2,3$) generalizes better than a single classifier, the misclassification of an unlabeled example is inevitable. Therefore, $h_{i}^{d}$ ($i=1,2,3$) will receive noisy examples as its source and target data, which might bias the refinement of $h_{i}^{d}$. Fortunately, Zhou et al. \citep{11} has proposed that the negative influence caused by such noise could be compensated by augmenting the labeled set with sufficient amount of newly labeled examples under certain conditions. Thus, being inspired from Zhou et al. \citep{11}, the same strategy is taken to restrict the influence caused by noise newly labeled samples in two cases. 

The first case is how to select unlabeled data in $U[d]$ to be labeled as target data for retraining $h_{i}^{d}$ ($i$=$1,2,3$). Let $L_{i}^{d,t}$ and $L_{i}^{d,t-1}$ denote the set of samples that are labeled for retraining $h_{i}^{d}$ in the $t$-$th$ round of iteration and $(t$-$1)$-$th$ round of iteration whose size are $|L_{i}^{d,t}|$ and $|L_{i}^{d,t-1}|$, respectively. $e_{i}^{d,t}$ and $e_{i}^{d,t-1}$ denote the upper boundary of the classification error rate of $H_{i}^{d}=h_{j}^{d}\cup h_{k}^{d}$ in the $t$-$th$ round of iteration and $(t$-$1)$-$th$ round of iteration, respectively. Therefore, according to Zhou et al. \citep{11}, the following constraint is obtained.
\begin{equation}
\label{newequal1}
0<\frac{e_{i}^{d,t}}{e_{i}^{d,t-1}}<\frac{|L_{i}^{d,t-1}|}{|L_{i}^{d,t}|}<1
\end{equation}
According to Eq.\ref{newequal1}, when $e_{i}^{d,t}<e_{i}^{d,t-1}$ and $|L_{i}^{d,t}|>|L_{i}^{d,t-1}|$ are met,  $e_{i}^{d,t}|L_{i}^{d,t}|<e_{i}^{d,t-1}|L_{i}^{d,t-1}|$ might still be violated since $L_{i}^{d,t}\gg L_{i}^{d,t-1}$ may occur. To make Eq.\ref{newequal1} hold again in this case, $L_{i}^{d,t}$ must be subsampled so that  $|L_{i}^{d,t}|<e_{i}^{d,t-1}|L_{i}^{d,t-1}|/e_{i}^{d,t}$. Therefore, $L_{i}^{d,t}$ should be randomly subsampled using following function:
\begin{equation}
\label{newequal3}
L_{i}^{d,t}\leftarrow Subsample(L_{i}^{d,t},\lceil \frac{e_{i}^{d,t-1}|L_{i}^{d,t-1}|}{e_{i}^{d,t}} - 1 \rceil)
\end{equation}

The second case is how to select unlabeled data in $U[(d+1)\%2]$ to be labeled as source data for retraining $h_{i}^{d}$ ($i$=$1,2,3$). Let $L^{d,t}$ and $L^{d,t-1}$ denote the set of samples that are labeled in the $t$-$th$ round of iteration and $(t$-$1)$-$th$ round of iteration whose size are $|L^{d,t}|$ and $|L^{d,t-1}|$, respectively. $e^{d,t}$ and $e^{d,t-1}$ denote the upper boundary of the classification error rate of $H^{d}$ in the $t$-$th$ round of iteration and $(t$-$1)$-$th$ round of iteration, respectively. Therefore, we have the following constraints:
\begin{equation}
\label{newequal2}
0<\frac{e^{d,t}}{e^{d,t-1}}<\frac{|L^{d,t-1}|}{|L^{d,t}|}<1
\end{equation}
According to Eq.\ref{newequal2}, when $e^{d,t}<e^{d,t-1}$ and $|L^{d,t}|>|L^{d,t-1}|$, is met, $e^{d,t}|L^{d,t}|<e^{d,t-1}|L^{d,t-1}|$  might still be violated since $L^{d,t}\gg L^{d,t-1}$ may occur. To make Eq.\ref{newequal2} hold again in this case, $L^{d,t}$ must be subsampled so that $|L^{d,t}|<e^{d,t-1}|L^{d,t-1}|/e^{d,t}$. Therefore, $L^{d,t}$ should be randomly subsampled using the following function:

\begin{equation}
\label{newequal4}
L^{d,t}\leftarrow Subsample(L^{d,t},\lceil \frac{e^{d,t-1}|L^{d,t-1}|}{e^{d,t}} - 1 \rceil)
\end{equation}

The constraint introduced in the second case brings more reliably pseudo-labeled samples to the source domain, makes the source domain become more accurate, and then leads to more reliably pseudo-labeled samples added to the target domain. This is very important for the iteration process in Co-Transfer.

\begin{algorithm}
  \caption{\emph{Co-Transfer}}
  \label{alg:Tri-evolving_1}
  \KwIn{$D_{SL};\ D_{SU};\ D_{TL};\ D_{TU};\ D;\ N$\\}
  \KwOut{$H^{1}(x)\leftarrow argmax_{y \in \mathcal{Y}} \sum_{i:h_{i}^{1}(x)=y}  1$}
  $L=[D_{TL},D_{SL}];U=[D_{TU},D_{SU}]$\;
  \For{$d \in \{0,1\}$}
  {
    $H^d\leftarrow \emptyset$\;
    \For{$i \in \{1,2,3\}$}
    {
      $S_{i}^{d}\leftarrow BootstrapSample(L[(d+1)\%2])$\;
      $T_{i}^{d}\leftarrow BootstrapSample(L[d])$\;
      $h_{i}^{d}\leftarrow TrAdaBoost(S_{i}^{d}, T_{i}^{d}, N, D)$\;
      $H^d\leftarrow H^d \cup h_{i}^{d}; e_{i}^{d'}\leftarrow 0.5; l_{i}^{d'} \leftarrow 0$\;
    }
    $e^{d'} \leftarrow 0.5; l^{d'} \leftarrow 0$\;
  }
\Repeat{$none\ of\ H^{d}(d \in {0,1})\ changes$}
  {
    \For{$d \in \{0,1\}$}
    {
      $L^{d}\leftarrow \emptyset; Update^d \leftarrow FALSE$\;
      $e^d=MeasureEnsembleError(H^{d},L[d])$\;
      \For{$i \in \{1,2,3\}$}
      {
        $L_{i}^{d}\leftarrow \emptyset; Update_{i}^{d} \leftarrow FALSE$\;
        $e_{i}^{d}=MeasureClassifierError(h_{j}^{d}\& h_{k}^{d}, L[d])$\;
        \If{$e_{i}^{d}<e_{i}^{d'}$}
        {
          $L_{i}^{d}\leftarrow PseudoLabel(h_{j}^{d}\& h_{k}^{d}, U[d])$\;
          \If{$l_{i}^{d'}<|L_{i}^{d}|$}
          {
            \If{$e_{i}^{d}|L_{i}^{d}|<e_{i}^{d'}l_{i}^{d'}$}
            {
              $Update_{i}^{d}=TRUE$\;
            }
            \ElseIf{$l_{i}^{d'}>\frac{e_{i}^{d}}{e_{i}^{d'}-e_{i}^{d}}$}
            {
              $L_{i}^{d}\leftarrow Subsample(L_{i}^{d},\lceil \frac{e_{i}^{d'}l_{i}^{d'}}{e_{i}^{d}} - 1 \rceil)$ \%refer Eq.\ref{newequal3}\;
              $Update_{i}^{d}=TRUE$\;
            }
          }
        }
      }
      \If{$e^d<e^{d'}$}
      {
        $L^{d}\leftarrow screenPseudoLabel(H^d, L_{i}^{d}, L_{j}^{d}, L_{k}^{d})$\;
        \If{$l^{d'}<|L^d|$}
        {
          \If{$e^{d}|L^{d}|<e^{d'}l^{d'}$}
          {
            $Update^{d}=TRUE$\;
          }
          \ElseIf{$l^{d'}>\frac{e^{d}}{e^{d'}-e^{d}}$}
          {
            $L^{d}\leftarrow Subsample(L^{d},\lceil \frac{e^{d'}l^{d'}}{e^{d}} - 1 \rceil)$ \%refer Eq.\ref{newequal4}\; 
            $Update^{d}=TRUE$\;
          }
        }
      }
    }
    \For{$d \in \{0,1\}$}
    {
      \For{$i \in \{1,2,3\}$}
      {
        \If{$Update_{i}^{d} \cap Update^{(d+1)\%2}$}
        {
          $NS_{i}^{d}\leftarrow L[(d+1)\%2] \cup L^{(d+1)\%2}$\;
          $NT_{i}^{d}\leftarrow L[d] \cup L_{i}^{d}$\;
          $Nh_{i}^{d}\leftarrow TrAdaBoost(NS_{i}^{d},NT_{i}^{d}, N, D)$\;
          $h_{i}^{d}\leftarrow Nh_{i}^{d}; e_{i}^{d'}\leftarrow e_{i}^{d};l_{i}^{d'}\leftarrow |L_{i}^{d}|$\;
        }
      }
      $e^{d'}\leftarrow e^d;l^{d'}\leftarrow |L^{d'}|$\;
    }
  }
\end{algorithm}

\begin{figure*}[htp]
\centering
\includegraphics[scale=0.75]{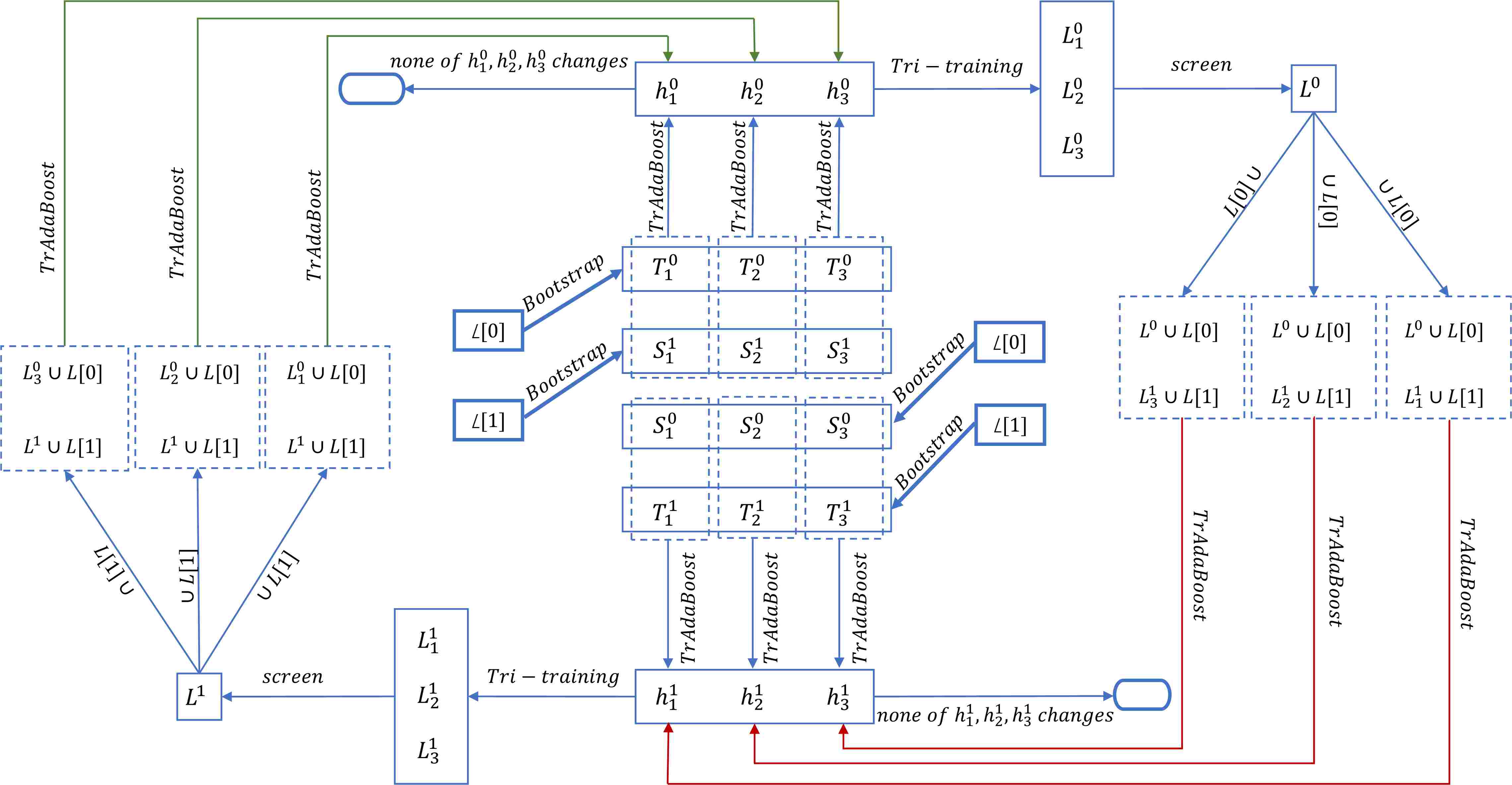}
\caption{The data flow diagram of \emph{Co-Transfer} training process}
\label{fig:framework}
\end{figure*}

The pseudo-code of Co-Transfer is presented in Algorithm \ref{alg:Tri-evolving_1}. Lines \emph{5-7} denote that $S_{i}^{d}$ and $T_{i}^{d}$ are obtained by $BootstrapSample$ from the original labeled data in $L[(d+1)\%2d]$ and $L[d]$ as the source and target data, respectively, and then a TrAdaBoost classifier is trained. In line \emph{13}, the function $MeasureEnsembleError$ attempts to estimate the classification error rate of $H^d$. In line \emph{16}, $MeasureClassifierError$ attempts to estimate the classification error rate of the hypothesis derived from the combination of $h_{j}^{d}$ and $h_{k}^{d}$. Here, both error rates are estimated on $L[d]$. In detail, the error rate of $H^d$ is approximated through dividing the number of labeled samples on which the three classifiers $h_{i}^{d}(i=1,2,3)$ of $H^d$ make incorrect classification by the number of labeled samples on which the classification made by $h_{i}^{d}=h_{j}^{d}=h_{k}^{d}$. Similarly, the error rate of the hypothesis of $h_{j}^{d}\& h_{k}^{d}$ is approximated through dividing the number of labeled samples on which both $h_{j}^{d}$ and $h_{k}^{d}$ make incorrect classification by the number of labeled samples on which the classification made by $h_{j}^{d}$ is the same as that made by $h_{k}^{d}$. In line \emph{18}, the function $PseudoLabel$ is to label the unlabeled samples in $U[d]$ when $h_{j}^{d}$ and $h_{k}^{d}$ agree for labeling. In line \emph{26}, the function $screenPseudoLabel$ is to screen the pseudo-labeled samples that are consistent with the ensemble $H^d$ from $L_{i}^{d}$, $L_{j}^{d}$, and $L_{k}^{d}$. In lines \emph{23} and \emph{31}, the function $Subsample(s,t)$ randomly removes $t$ number of examples from $s$. Lines \emph{33-40} employ the newly labeled source and target data to refine each base classifier of $H^d$. The data flow diagram during the training process of Co-Transfer is shown in Figure \ref{fig:framework}.

Co-Transfer combines instance-based transfer learning and semi-supervised learning to fully explore the knowledge in the existing data, including the source domain knowledge that is different from the target domain distribution but related, and the knowledge of the unlabeled samples in the source and target domains.

\section{Experiments}
\subsection{Datasets}
Five UCI\footnote{http://www.ics.uci.edu/˜mlearn/MLRepository.html} data sets and the Reuters\footnote{http://www.daviddlewis.com/resources/testcollections/} text classification task data set are used in the experiments. Table \ref{tab:exp_dataset} tabulates the detailed information of these data sets, where ``attr" represents the number of attributes, ``$|D_S|$" and ``$|D_T|$" represents the size of the source domain and the target domain, respectively, ``class" represents the number of labels of data, ``$h^s\rightarrow D_T$" represents the prediction error rate on $D_T$ made by the classifier trained by the source domain data, while  ``$h^t\rightarrow D_S$" represents the prediction error rate on $D_S$ made by the classifier trained by the target domain data. $h^s\rightarrow D_T$ is employed to evaluate the transferability from the source domain to the target domain, while $h^t\rightarrow D_S$ is used to evaluate the transferability from the target domain to the source domain. Such operations can be found in \citep{24}.

\begin{table}[htbp]
	\caption{\label{tab:test}EXPERIMENTAL DATA SETS} 
     \label{tab:exp_dataset}{
	\begin{center}{
		\begin{tabular}{m{3cm}ccccccc}
			\toprule
			Date set & attr & $|D_S|$ & $|D_T|$ & class & \emph{$h^s\rightarrow D_T$} & \emph{$h^t\rightarrow D_S$}\\
			\midrule
			Mushroom & 22 & 4608 & 3516 & 2 & 0.332 & 0.646\\
			Waveform & 21 & 1722 & 1582 & 2 & 0.171 & 0.135\\
			Magic & 10 & 9718 & 9302 & 2 & 0.332 & 0.377\\
			Splice & 60 & 795 & 740 & 2 & 0.116 & 0.085\\
			Vote & 16 & 220 & 215 & 2 & 0.079 & 0.118\\
			Orgs vs People & 4771 & 1237 & 1208 & 2 & 0.408 & 0.335\\
			Orgs vs Places & 4415 & 1016 & 1043 & 2 & 0.392 & 0.365\\
			People vs Places & 4562 & 1077 & 1077 & 2 & 0.463 & 0.443\\
			\bottomrule 
		\end{tabular}
		}
	\end{center}
	}
\end{table}

The five UCI data sets are first preprocessed for semi-supervised transfer learning. Among them, for the Mushroom, Waveform, Magic, and Splice data sets, the preprocessing method is the same as the method taken in the paper \citep{24}. These data sets have been proved that the source domain could assist the target domain task to better learn the hypothesis, that is, there is a positive transfer. To verify the effectiveness of the Co-Transfer algorithm on small-scale data set, the Vote data set is introduced. For each instance in the Vote data set, if its fourth attribute value is 0, it is put into the target domain $D_T$; otherwise, the instance is added to the source domain $D_S$. The preprocessed Reuters text classification data set is used to evaluate the classification accuracy of Co-Transfer on real application task. It can be downloaded directly, which includes three data sets, namely \emph{Orgs vs People}, \emph{Orgs vs Places} and \emph{People vs places}.

\subsection{Experimental settings}
For each data set, \emph{five-fold} cross validation is employed for reliable evaluation. In each fold, the data in the target domain are divided into the target domain training data set $D_T$ and the test data set $D_{Test}$. Then, the source domain data set and the target training data set are randomly partitioned into two labeled sets $D_{SL}$ and $D_{TL}$ and two unlabeled set $D_{SU}$ and $D_{TU}$, respectively, with a given label rate, i.e., 10\%, 20\%, 40\%, and 50\%. In addition, in each fold of cross validation, the source domain training data set is randomly partitioned into $D_{SL}$ and $D_{SU}$ twice, and the target domain training data set is randomly partitioned into $D_{TL}$ and $D_{TU}$ three times. Therefore, the final error rate is the average of 30=(5*6) test results.

To verify Co-Transfer could not only benefit from the labeled data in the source and target domains but also from the unlabeled data in the source and target domains, four baseline methods are introduced for comparison. Table \ref{tab:baseline} lists the data strategy used by Co-Transfer and the baseline methods, including \emph{DT}, \emph{TrAdaBoost}, \emph{Tri-training} and \emph{$TrAdaBoost_A$}. \emph{DT}, which is trained only on $D_{TL}$, is introduced as the worst classifier for comparison, while \emph{$TrAdaBoost_A$}, which is trained on $D_{SL}$, $D_{SU}$, $D_{TL}$, and $D_{TU}$ provided with all ground-truth labels of all unlabeled data, is introduced as the best classifier for comparison. \emph{TrAdaBoost}, which is trained on $D_{SL}$ and $D_{TL}$, is introduced for comparison to illustrate Co-Transfer can benefit from unlabeled samples in addition to reuse the source domain data. \emph{Tri-training}, which is trained on $D_{TL}$ and $D_{TU}$, is introduced for comparison to show Co-Transfer could benefit from the source domain data.

\begin{table}[htbp]
	\caption{The data strategy used by the algorithms} 
	\label{tab:baseline}
	\begin{center}{
		\begin{tabular}{cccccc}
			\toprule
			\multicolumn{1}{c}{\multirow{2}{*}{\emph{Baseline}}} & \multicolumn{4}{c}{\emph{Training Data}} & \multicolumn{1}{c}{\multirow{2}{*}{\emph{Test Data}}}\\
			\cline{2-5}
			 & $D_{SL}$ & $D_{SU}$ & $D_{TL}$ & $D_{TU}$ & \\
			
			\midrule
			\emph{$DT$} & $/$ & $/$ & $\surd$ & $/$ & $D_{Test}$\\
			\emph{$TrAdaBoost$} & $\surd$ & $/$ & $\surd$ & $/$ & $D_{Test}$\\
			\emph{$Tri$-$training$} & $/$ & $/$ & $\surd$ & $\times$ & $D_{Test}$\\
			\emph{$Co$-$Transfer$} & $\surd$ & $\times$ & $\surd$ & $\times$ & $D_{Test}$\\
			\emph{$TrAdaBoost_{A}$} & $\surd$ & $\surd$ & $\surd$ & $\surd$ & $D_{Test}$\\
			\bottomrule 
		\end{tabular} 
		}
    \end{center}
\end{table}

Since the DT and Tri-training algorithms do not require hyperparameters, they can be run directly. For TrAdaBoost, Co-Transfer and $TrAdaBoost_A$, the grid search is used to find suitable parameters. As a result, for UCI data sets, $N$=10 and $D$=10 are used in Mushroom, $N$=65 and $D$=4 in  Waveform, $N$=35 and $D$=20 in Magic, $N$=15 and $D$=50 in Splice, and $N$=5 and $D$=50 in Vote. For the text classification data sets, $N$=60 and $D$=4 are used in \emph{Orgs vs People}; $N$=30; $D$=4 in \emph{Orgs vs Places}; $N$=60; $D$=3 in \emph{People vs Places}.

\subsection{Experimental results and analysis}
Tables \ref{tab:10label}-\ref{tab:50label} tabulate the error rates of Co-Transfer and the baseline methods under different label rates. The significance is checked using a standard t-test with 95\% confidence between Co-Transfer and each baseline method. The $\bullet$/$\circ$ indicates that Co-Transfer is significantly better or worse than the corresponding baseline method, while the $\star$ means that Co-Transfer is not significantly better than the compared baseline method with 95\% confidence. The row \emph{avg.} shows the average results over all the experimental data sets.

\begin{table}[htbp]
	\caption{The error rates of the compared algorithms under the label rate of 10\%} 
	\label{tab:10label}
	\begin{center}{
		\resizebox{\textwidth}{0.8in}{
			\begin{tabular}{cccccccccc}
				\toprule
				\multicolumn{1}{c}{\multirow{2}{*}{\emph{DataSet}}} & \multicolumn{1}{c}{\multirow{2}{*}{\emph{$DT$}}} & \multicolumn{1}{c}{\multirow{2}{*}{\emph{$TrAdaBoost$}}} & \multicolumn{3}{c}{\emph{$Tri$-$training$}} & \multicolumn{3}{c}{\emph{$Co$-$Transfer$}} & \multicolumn{1}{c}{\multirow{2}{*}{\emph{$TrAdaBoost_{A}$}}}\\
				\cline{4-9}
				 ~&~ & ~& {Initial} & \emph{Final} & \emph{Iter} & \emph{Initial} & \emph{Final} & \emph{Iter} & ~ \\
				
				\midrule
				Mushroom & 0.004 $\star$ & 0.006 $\star$ & 0.008 & 0.004 $\star$ & 2.4 & 0.118 & 0.005 & 3 & \textbf{0.0} $\circ$\\
				Waveform & 0.201 $\bullet$ & 0.149 $\bullet$ & 0.193 & 0.19 $\bullet$ & 3 & 0.179 & \textbf{0.137} & 3 & 0.158 $\bullet$\\
				Magic & 0.155 $\bullet$ & 0.159 $\bullet$ & 0.161 & 0.149 $\star$ & 3 & 0.157 & 0.148 & 3.43 & \textbf{0.113} $\circ$\\
				Splice & 0.135 $\bullet$ & 0.122 $\bullet$ & 0.132 & 0.112 $\star$ & 2.8 & 0.16 & 0.111 & 3 & \textbf{0.055} $\circ$\\
				Vote & 0.054 $\star$ & 0.045 $\star$ & 0.064 & 0.051 $\star$ & 2.27 & 0.086 & 0.044 & 2.87 & \textbf{0.02} $\circ$\\
				Orgs vs People & 0.264 $\bullet$ & 0.215 $\bullet$ & 0.298 & 0.258 $\bullet$ & 3 & 0.377 & 0.191 & 3.07 & \textbf{0.153} $\circ$\\
				Orgs vs Places & 0.282 $\bullet$ & 0.264 $\bullet$ & 0.307 & 0.274 $\bullet$ & 3 & 0.384 & 0.234 & 3.17 & \textbf{0.179} $\circ$\\
				People vs Places & 0.274 $\bullet$ & 0.227 $\bullet$ & 0.303 & 0.264 $\bullet$ & 3 & 0.417 & 0.206 & 3 & \textbf{0.159} $\circ$\\
				avg. & 0.171 & 0.148 & 0.183 & 0.163 & 2.81 & 0.235 & 0.135 & 3.07 & \textbf{0.105}\\
				\bottomrule 
			\end{tabular} 
			}
		}
    \end{center}
\end{table}

\begin{table}[htbp]
	\caption{The error rates of the compared algorithms under the label rate of 20\%} 
	\label{tab:20label}
	\begin{center}{
		\resizebox{\textwidth}{0.8in}{
			\begin{tabular}{cccccccccc}
				\toprule
				\multicolumn{1}{c}{\multirow{2}{*}{\emph{DataSet}}} & \multicolumn{1}{c}{\multirow{2}{*}{\emph{$DT$}}} & \multicolumn{1}{c}{\multirow{2}{*}{\emph{$TrAdaBoost$}}} & \multicolumn{3}{c}{\emph{$Tri$-$training$}} & \multicolumn{3}{c}{\emph{$Co$-$Transfer$}} & \multicolumn{1}{c}{\multirow{2}{*}{\emph{$TrAdaBoost_{A}$}}}\\
				\cline{4-9}
				 ~&~ & ~& \emph{Initial} & \emph{Final} & \emph{Iter} & \emph{Initial} & \emph{Final} & \emph{Iter} & ~ \\
				
				\midrule
				Mushroom & 0.002 $\star$ & 0.001 $\star$ & 0.003 & 0.002 $\star$ & 2.2 & 0.081 & 0.002 & 3 & \textbf{0.0} $\circ$\\
				Waveform & 0.179 $\bullet$ & 0.146 $\bullet$ & 0.17 & 0.169 $\bullet$ & 3 & 0.159 & \textbf{0.127} & 3 & 0.159 $\bullet$\\
				Magic & 0.149 $\bullet$ & 0.147 $\bullet$ & 0.146 & 0.139 $\bullet$ & 3 & 0.147 & 0.127 & 3.53 & \textbf{0.116} $\circ$\\
				Splice & 0.103 $\star$ & 0.095 $\star$ & 0.101 & 0.095 $\star$ & 2.93 & 0.113 & 0.092 & 3 & \textbf{0.054} $\circ$\\
				Vote & 0.047 $\star$ & 0.036 $\star$ & 0.047 & 0.039 $\star$ & 2.2 & 0.045 & 0.039 & 3.07 & \textbf{0.018} $\circ$\\
				Orgs vs People & 0.213 $\bullet$ & 0.182 $\bullet$ & 0.22 & 0.198 $\bullet$ & 3 & 0.353 & 0.155 & 3.27 & \textbf{0.15} $\star$\\
				Orgs vs Places & 0.242 $\bullet$ & 0.238 $\bullet$ & 0.252 & 0.231 $\bullet$ & 3 & 0.365 & 0.205 & 3.6 & \textbf{0.186} $\circ$\\
				People vs Places & 0.217 $\bullet$ & 0.201 $\bullet$ & 0.249 & 0.199 $\bullet$ & 3 & 0.42 & 0.17 & 3.43 & \textbf{0.155} $\star$\\
				avg. & 0.144 & 0.131 & 0.149 & 0.134 & 2.79 & 0.21 & 0.115 & 3.24 & \textbf{0.105}\\
				\bottomrule 
			\end{tabular} 
			}
		}
    \end{center}
\end{table}

\begin{table}[htbp]
	\caption{The error rates of the compared algorithms under the label rate of 40\%} 
	\label{tab:40label}
	\begin{center}{
		\resizebox{\textwidth}{0.8in}{
			\begin{tabular}{cccccccccc}
				\toprule
				\multicolumn{1}{c}{\multirow{2}{*}{\emph{DataSet}}} & \multicolumn{1}{c}{\multirow{2}{*}{\emph{$DT$}}} & \multicolumn{1}{c}{\multirow{2}{*}{\emph{$TrAdaBoost$}}} & \multicolumn{3}{c}{\emph{$Tri$-$training$}} & \multicolumn{3}{c}{\emph{$Co$-$Transfer$}} & \multicolumn{1}{c}{\multirow{2}{*}{\emph{$TrAdaBoost_{A}$}}}\\
				\cline{4-9}
				 ~&~ & ~& \emph{Initial} & \emph{Final} & \emph{Iter} & \emph{Initial} & \emph{Final} & \emph{Iter} & ~ \\
				
				\midrule
				Mushroom & 0.0 $\star$ & 0.0 $\star$ & 0.001 & 0.0 $\star$ & 2.07 & 0.051 & 0.0 & 3 & 0.0 $\star$\\
				Waveform & 0.174 $\bullet$ & 0.144 $\bullet$ & 0.154 & 0.165 $\bullet$ & 3 & 0.157 & \textbf{0.135} & 3 & 0.158 $\bullet$\\
				Magic & 0.144 $\bullet$ & 0.134 $\bullet$ & 0.146 & 0.13 $\bullet$ & 3 & 0.136 & 0.113 & 3.37 & \textbf{0.109} $\star$\\
				Splice & 0.085 $\star$ & 0.09 $\star$ & 0.072 & 0.07 $\star$ & 3 & 0.089 & 0.084 & 3 & \textbf{0.054} $\circ$\\
				Vote & 0.029 $\star$ & 0.025 $\star$ & 0.031 & 0.025 $\star$ & 2.47 & 0.027 & 0.023 & 3 & \textbf{0.016} $\circ$\\
				Orgs vs People & 0.18 $\bullet$ & 0.194 $\bullet$ & 0.177 & 0.165 $\bullet$ & 3 & 0.311 & \textbf{0.147} & 3.8 & 0.154 $\star$\\
				Orgs vs Places & 0.197 $\bullet$ & 0.225 $\bullet$ & 0.222 & 0.19 $\bullet$ & 3 & 0.352 & \textbf{0.176} & 3.8 & 0.182 $\star$\\
				People vs Places & 0.19 $\bullet$ & 0.19 $\bullet$ & 0.195 & 0.177 $\bullet$ & 3 & 0.38 & \textbf{0.142} & 3.67 & 0.155 $\star$\\
				avg. & 0.125 & 0.125 & 0.125 & 0.115 & 2.82 & 0.188 & \textbf{0.102} & 3.33 & 0.104\\
				\bottomrule 
			\end{tabular} 
			}
		}
    \end{center}
\end{table}

\begin{table}[htbp]
	\caption{The error rates of the compared algorithms under the label rate of 50\%} 
	\label{tab:50label}
	\begin{center}{
		\resizebox{\textwidth}{0.8in}{
			\begin{tabular}{cccccccccc}
				\toprule
				\multicolumn{1}{c}{\multirow{2}{*}{\emph{DataSet}}} & \multicolumn{1}{c}{\multirow{2}{*}{\emph{$DT$}}} & \multicolumn{1}{c}{\multirow{2}{*}{\emph{$TrAdaBoost$}}} & \multicolumn{3}{c}{\emph{$Tri$-$training$}} & \multicolumn{3}{c}{\emph{$Co$-$Transfer$}} & \multicolumn{1}{c}{\multirow{2}{*}{\emph{$TrAdaBoost_{A}$}}}\\
				\cline{4-9}
				 ~&~ & ~& \emph{Initial} & \emph{Final} & \emph{Iter} & \emph{Initial} & \emph{Final} & \emph{Iter} & ~ \\
				
				\midrule
				Mushroom & 0.0 $\star$ & 0.0 $\star$ & 0.001 & 0.0 $\star$ & 2.13 & 0.04 & 0.0 & 3 & 0.0 $\star$\\
				Waveform & 0.176 $\bullet$ & 0.151 $\bullet$ & 0.159 & 0.167 $\bullet$ & 3 & 0.151 & \textbf{0.138} & 3.03 & 0.159 $\bullet$\\
				Magic & 0.143 $\bullet$ & 0.136 $\bullet$ & 0.15 & 0.13 $\bullet$ & 3 & 0.138 & \textbf{0.108} & 3.4 & 0.111 $\star$\\
				Splice & 0.082 $\star$ & 0.083 $\star$ & 0.076 & 0.069 $\star$ & 3 & 0.088 & 0.076 & 3 & \textbf{0.054} $\circ$\\
				Vote & 0.029 $\star$ & 0.027 $\star$ & 0.034 & 0.026 $\star$ & 2.53 & 0.029 & 0.026 & 3 & \textbf{0.016} $\star$\\
				Orgs vs People & 0.17 $\bullet$ & 0.2 $\bullet$ & 0.175 & 0.162 $\bullet$ & 3 & 0.313 & \textbf{0.142} & 4.03 & 0.152 $\star$\\
				Orgs vs Places & 0.187 $\bullet$ & 0.22 $\bullet$ & 0.188 & 0.168 $\star$ & 3 & 0.352 & \textbf{0.168} & 4 & 0.178 $\star$\\
				People vs Places & 0.183 $\bullet$ & 0.196 $\bullet$ & 0.181 & 0.166 $\bullet$ & 3 & 0.341 & \textbf{0.142} & 4.03 & 0.159 $\star$\\
				avg. & 0.121 & 0.127 & 0.12 & 0.111 & 2.83 & 0.182 & \textbf{0.1} & 3.44 & 0.104\\
				\bottomrule 
			\end{tabular} 
			}
		}
    \end{center}
\end{table}

In the columns of \emph{Tri-training} and  \emph{Co-Transfer}, the \emph{Initial} column shows the error rate of the hypothesis trained only on the original labeled data, which is $D_{TL}$ for \emph{Tri-training} and $D_{SL}$ and $D_{TL}$ for \emph{Co-Transfer}. The \emph{Final} column shows the error rate of the hypothesis being further refined with the pseudo-labeled data, which are from $D_{TU}$ for \emph{Tri-training} and $D_{SU}$ and $D_{TU}$ for \emph{Co-Transfer}. \emph{Iter} represents the number of learning rounds from \emph{Initial} to \emph{Final}. 

Tables \ref{tab:10label}-\ref{tab:50label} show that not only the unlabeled data in the source and target domains can be explored, but the knowledge of the source domain can also be reused to improve the accuracy of the hypothesis learned by Co-Transfer under different label rates. Under different label rates, the final hypothesis learned by Co-Transfer reaches the lowest average error rates except $TrAdaBoost_A$. However, it was surprising, when the average error rate of $TrAdaBoost_A$ is compared with that of Co-Transfer, it can be observed that the accuracy of the final hypothesis learned by Co-Transfer is comparable to that of the classifier learned by $TrAdaBoost_A$ when the label rate is set at 40\% and 50\%. 

To analyze Co-Transfer can benefit from unlabeled data, we compare Co-Transfer with TrAdaBoost which do not use unlabeled data to improve the accuracy of the learned hypothesis. From Tables \ref{tab:10label}-\ref{tab:50label}, it can be observed that the average error rate of the final hypothesis learned by Co-Transfer is lower that the classifier learned by TrAdaBoost under different label rates. In detail, by averaging on all the data sets, Co-Transfer achieves an average error rate of 0.135 and 0.115 under 10\% and 20\% label rates,  respectively. In addition, when the label rate increases to 40\% and 50\%, TrAdaBoost cannot further benefit from source domain samples. However, even in this case, Co-Transfer can still explore the unlabeled examples to achieve better performance than that of TrAdaBoost. When the label rate is 40\% and 50\%, Co-Transfer reaches an average error rate of 0.102 and 0.1 respectively.

In addition to benefiting from unlabeled examples, Co-Transfer can also reuse the knowledge of the source domain to further improve the accuracy of the learned hypothesis. To investigate the effectiveness of Co-Transfer, Tri-training, a well-known semi-supervised learning algorithm which do not benefit from source domain, is introduced for comparison. Since both Co-Transfer and Tri-training are iterative method, three different indicators are compared, including \emph{Initial}, \emph{Final} and \emph{Iter}. From Tables \ref{tab:10label}-\ref{tab:50label}, it can be observed that the error rate of the initial hypothesis learned by Co-Transfer is higher than that of the initial hypothesis learned by Tri-training on all the data sets under different label rates. However, the error rate of the final hypothesis learned by Co-Transfer is lower than that of Tri-training while the average number of iterations of Co-Transfer is slightly more than that of Tri-training by counting on all the data sets under all label rates. In detail, the average number of iterations of Co-Transfer is only 3.27 while that of Tri-training is 2.81. Moreover, the average error rate of the final hypothesis learned by Co-Transfer is reduced by 2.8\%, 1.9\%, 1.3\%, and 1.1\% on all experimental data sets, respectively, than those achieved by Tri-training under different label rates. These results can illustrate the iterative process of Co-Transfer is more effective than that of Tri-training.

To investigate the effectiveness of Co-Transfer in practical application scenarios, such as text classification tasks, the performance of Co-Transfer on the datasets of \emph{Orgs vs People}, \emph{Orgs vs Places} and \emph{People vs Places} are further analyzed. It can be observed from Tables \ref{tab:10label}-\ref{tab:50label} that Co-Transfer can enhance the accuracy of the learned hypothesis by exploring the knowledge of the source domain and the  unlabeled data under different label rates. By comparing Co-Transfer with TrAdaBoost and DT respectively, the error rate of the final hypothesis learned by Co-Transfer is reduced by 4.0\% and 4.9\% on \emph{Orgs vs People}, 4.2\% and 3.2\% on \emph{Orgs vs Places}, and 3.8\% and 5.1\% on \emph{People vs Places} by averaging four different label rates respectively. When comparing Co-Transfer and Tri-training, by averaging over four different label rates, such as 10\%, 20\%, 40\%, and 50\%, Co-Transfer achieves a reduction in error rate of 3.7\% on \emph{Orgs vs People}, 2.1\% on \emph{Orgs vs Places}, and 3.7\% on \emph{People vs Places}. This results show that Co-Transfer can not only explore the  unlabeled data but also reuse the knowledge of the source domain to improve the classification accuracy of the learned hypothesis. 

\begin{figure*}[htbp]
\centering
\subfigure[\emph{Orgs vs People}]{
\begin{minipage}[t]{0.33\linewidth}
\centering
\includegraphics[width=2in]{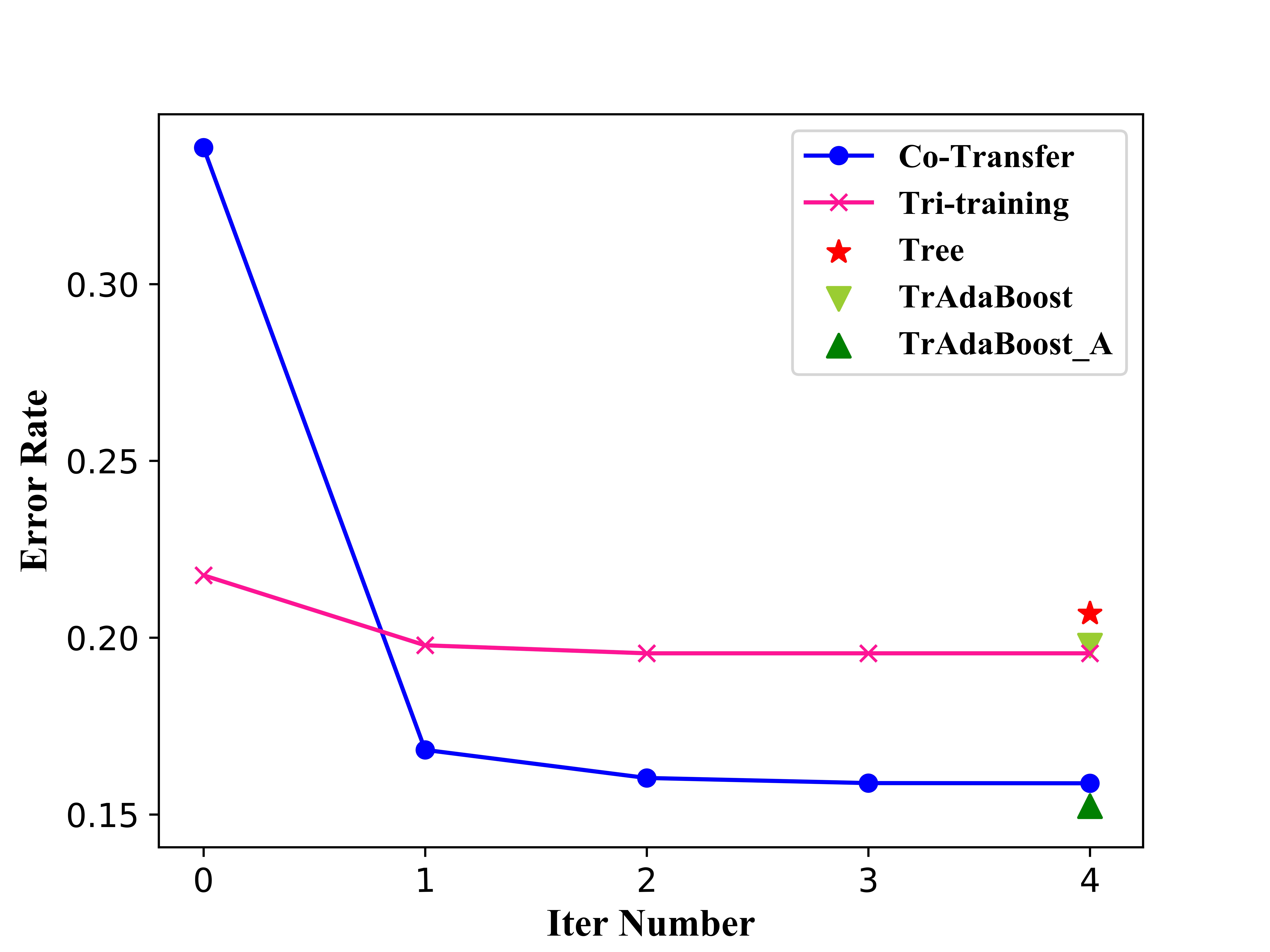}
\end{minipage}%
}%
\subfigure[\emph{Orgs vs Places}]{
\begin{minipage}[t]{0.33\linewidth}
\centering
\includegraphics[width=2in]{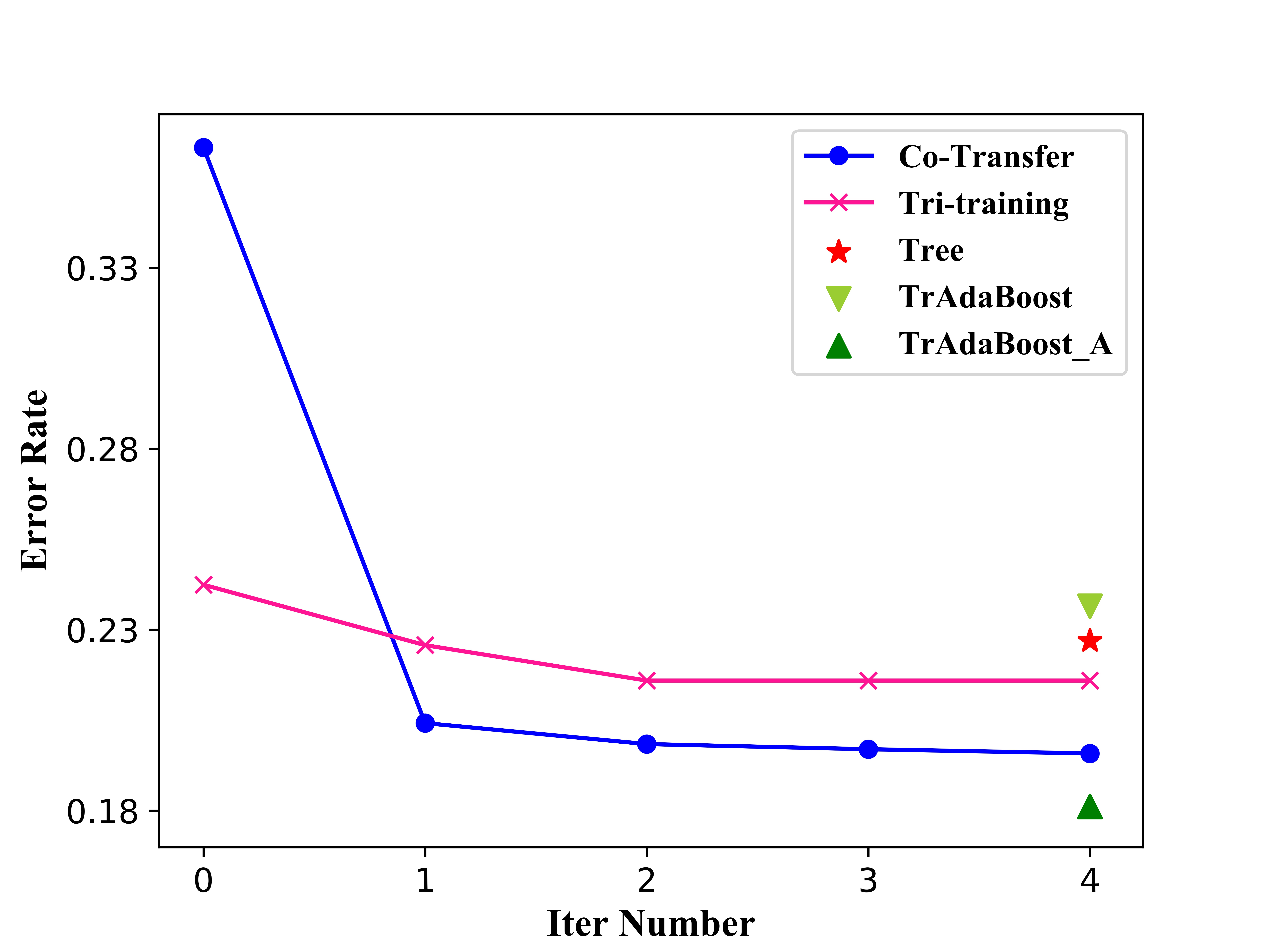}
\end{minipage}%
}%
\subfigure[\emph{People vs Places}]{
\begin{minipage}[t]{0.33\linewidth}
\centering
\includegraphics[width=2in]{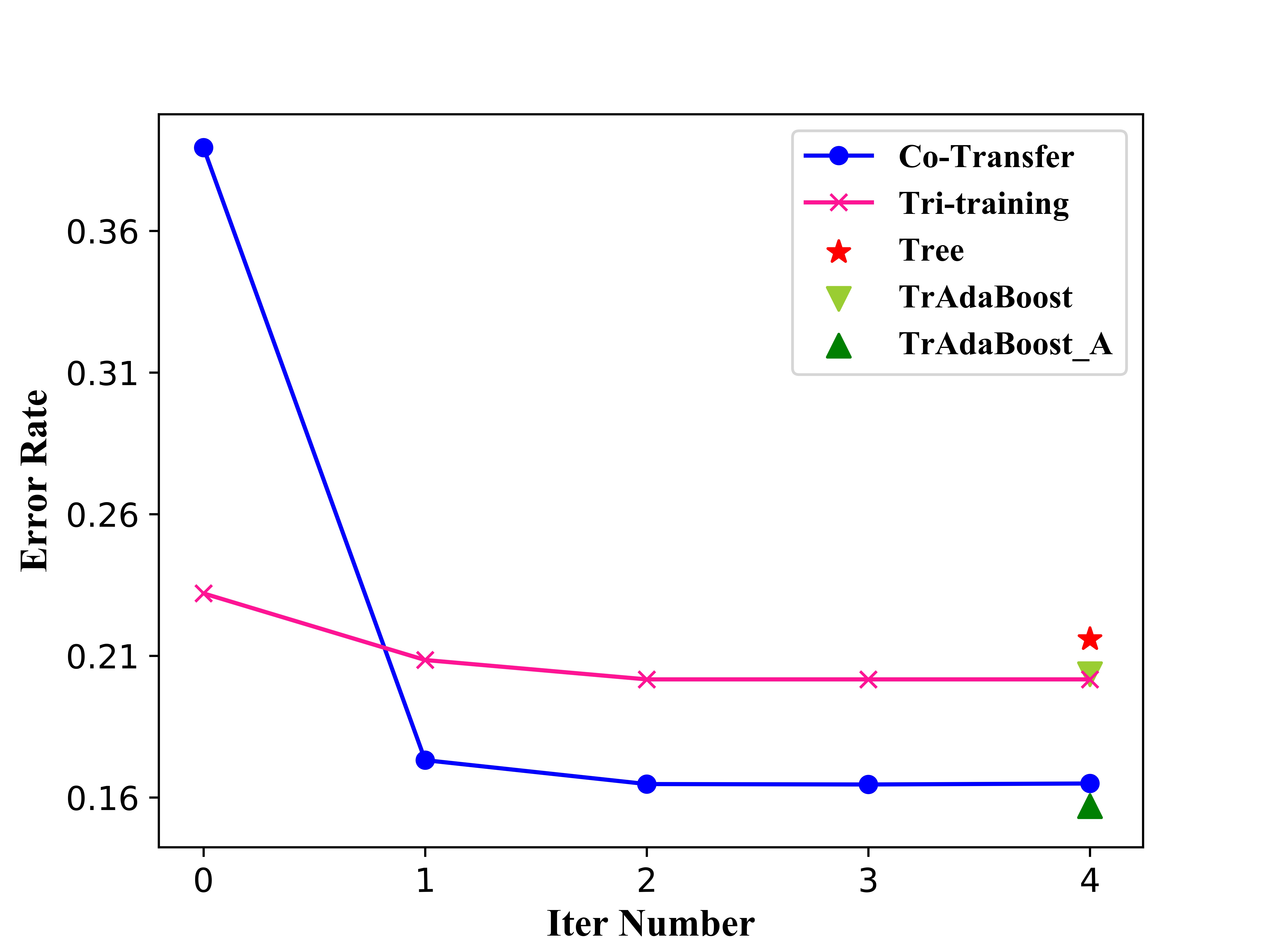}
\end{minipage}
}
\centering
\caption{The error rate averaged under different label rates over text classification task data sets}
\label{fig:error}
\end{figure*}

At last, to get an insight into the learning process of Co-Transfer, the error rates at each learning iteration are averaged over different label rates on the text classification task data sets, including \emph{Orgs vs People}, \emph{Orgs vs Places}, and \emph{People vs Places}. Note that among Co-Transfer and the baseline methods, only Tri-training and Co-Transfer can iteratively improve the learned hypothesis in training process, while the other three algorithms do not have an iterative learning process. Figure \ref{fig:error} depicts the changes in the average error rates of the compared algorithms from the initial iteration to the final iteration. It can be observed from Figure \ref{fig:error} that the initial average error rate of Co-Transfer is higher than the average error rate of the compared Tri-training algorithm. After the initial iteration, the lines of Co-Transfer are always below those of Tri-training, and the average error rates of Co-Transfer keep on decreasing after utilizing unlabeled data and source domain knowledge and quickly converges within just a few learning iterations. These advantages make Co-Transfer to greatly save the cost of manpower of labeling and reuse pre-existing examples to avoid the waste of knowledge.

\begin{figure}[htbp]
\centering
\subfigure[\emph{Orgs vs People}]{
\begin{minipage}[t]{0.33\textwidth}
\centering
\includegraphics[width=2in]{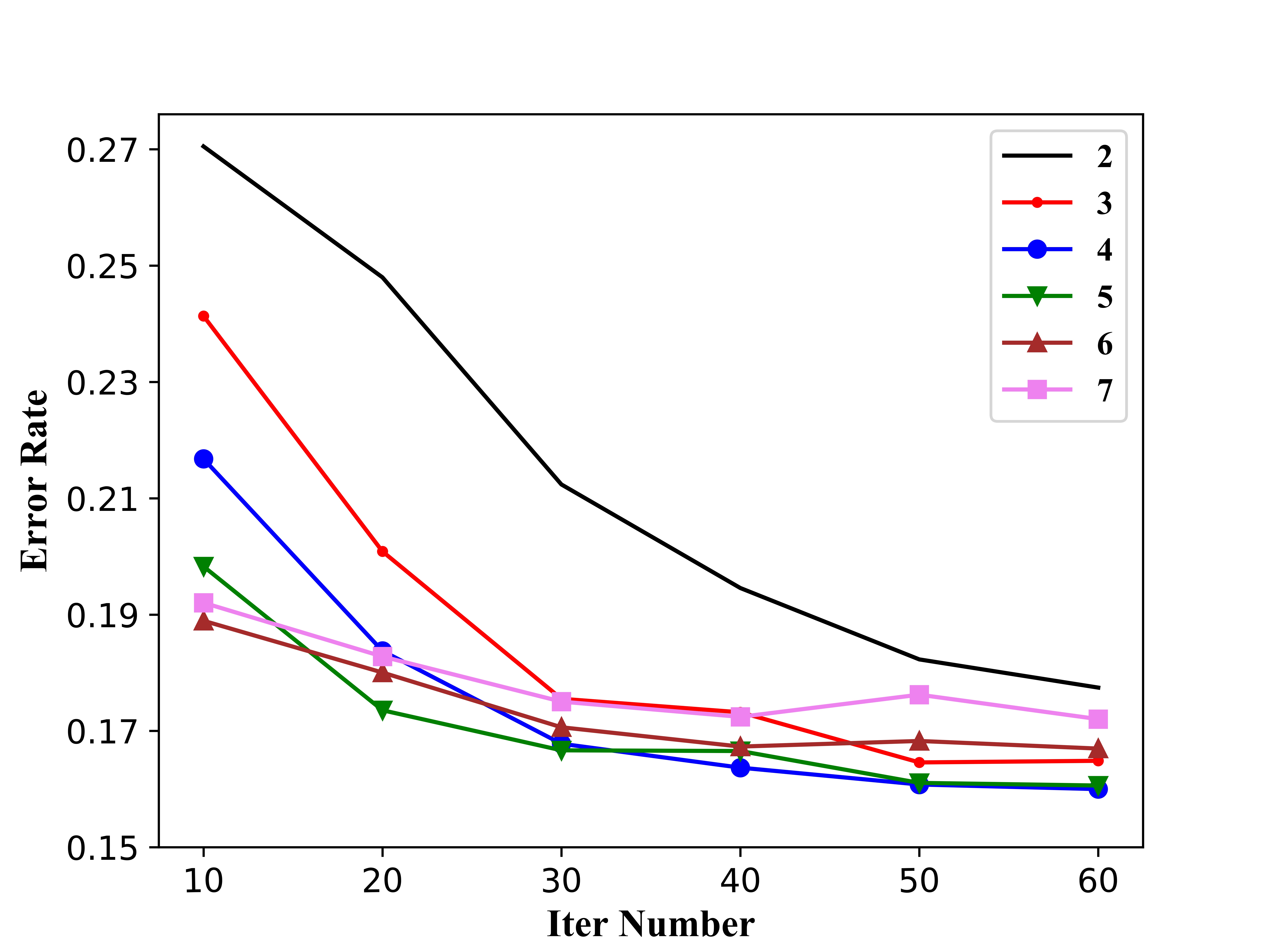}
\end{minipage}%
}%
\subfigure[\emph{Orgs vs Places}]{
\begin{minipage}[t]{0.33\linewidth}
\centering
\includegraphics[width=2in]{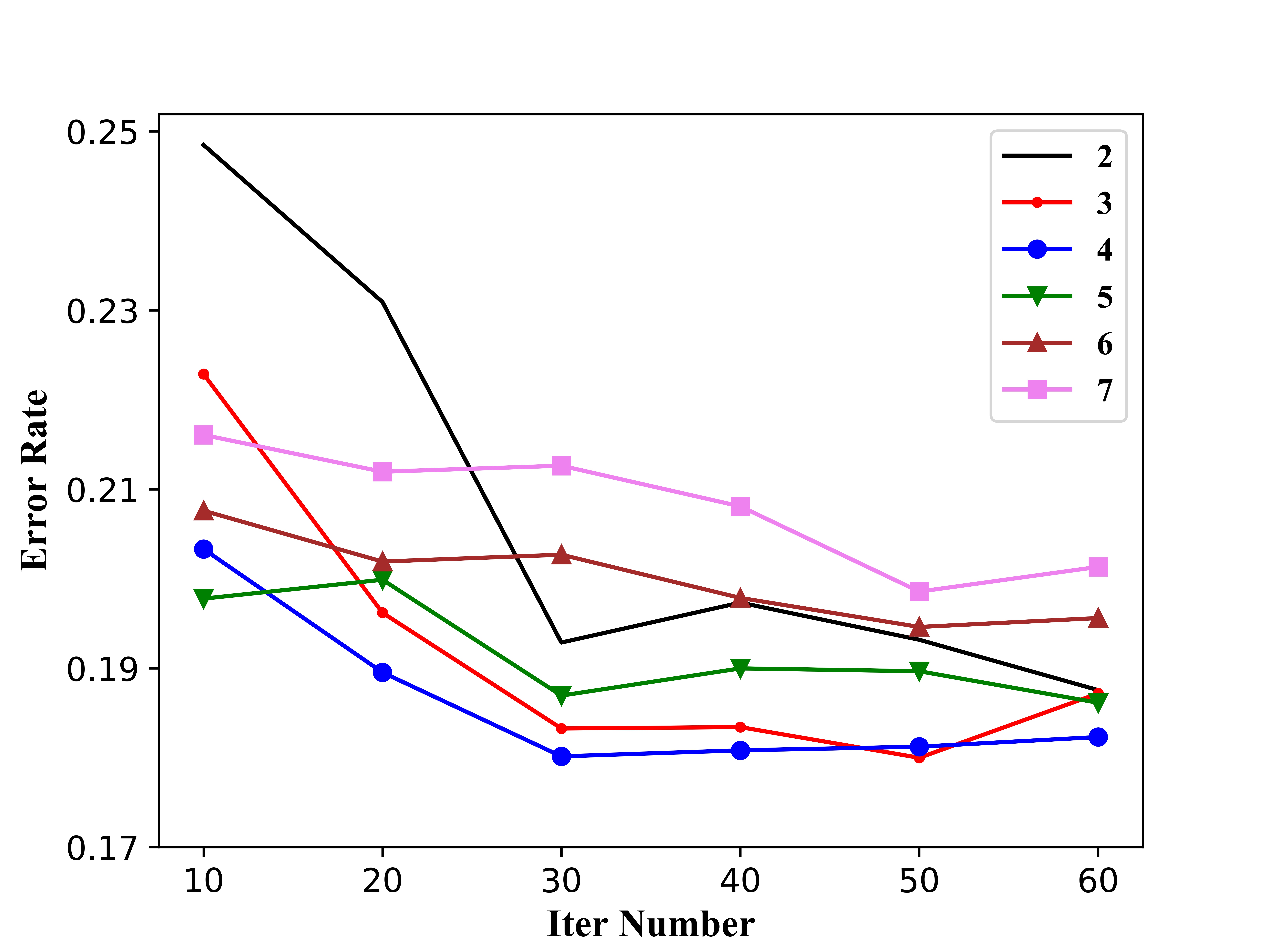}
\end{minipage}%
}%
\subfigure[\emph{People vs Places}]{
\begin{minipage}[t]{0.33\linewidth}
\centering
\includegraphics[width=2in]{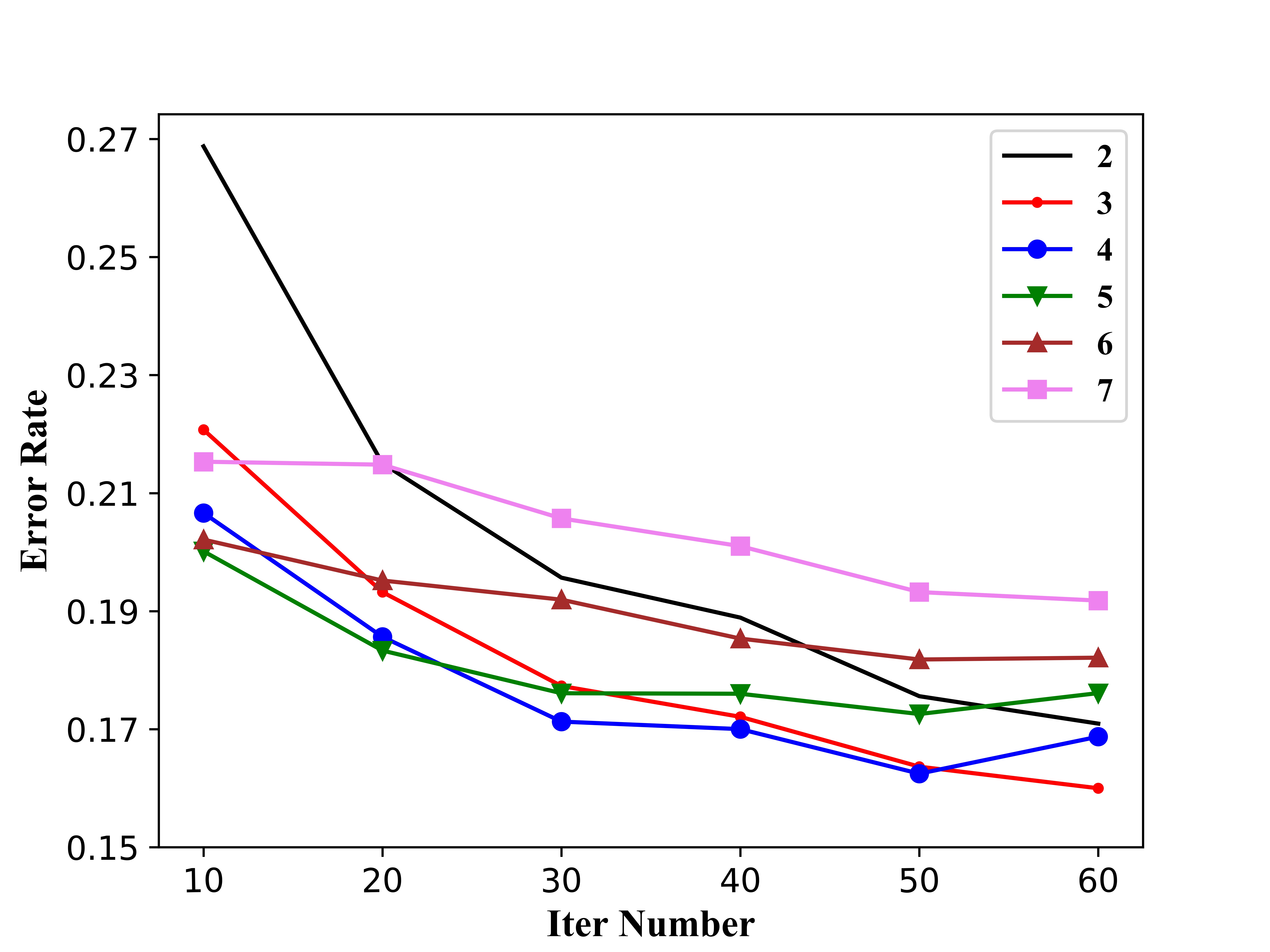}
\end{minipage}
}
\centering
\caption{The error rate is averaged over all label rates with different $N$ and $D$}
\label{fig:parameters}
\end{figure}

Note that in previous experiments, the depth $D$ of the tree and the number of iterations $N$ are fixed in Co-Transfer. Different $D$ and $N$ might affect the complexity of model and influence classification accuracy. Therefore, the error rates of Co-Transfer with different depth $D$ $(D=2,3,\dots,6,7)$ of the tree and the number of iterations $N$ $(N=10,20,\dots,50,60)$ are further investigated on the text classification task data sets. By averaging over all label rates, the variations of average error rate of Co-Transfer on the text classification data sets are shown in Figure \ref{fig:parameters}. In these figures, the lines with different color represent the changes of the average error rates with different depths $D$ of the tree under different iteration times $N$. It can be observed that the values of the parameters $N$ and $D$ for the optimal performance of Co-Transfer are different on the three data sets.  These observation illustrate that Co-Transfer is more sensitive to the parameter selection of the model for different data sets. However, it can be observed from Figure \ref{fig:parameters} that the error rate declines with the increase of $N$ on all datasets. 

\section{Conclusion}

In this paper, a semi-supervised inductive transfer learning framework Co-Transfer is proposed. Co-Transfer finely combine inductive transfer learning and semi-supervised learning , which can reuse the existing knowledge in source domain data and explore unlabeled data to boost the classification accuracy of the learned hypothesis. To ensure this, in Co-Transfer, two groups of three TrAdaBoost classifiers are employed to refine for producing the final hypothesis. In each round of iteration in Co-Transfer, each group of TrAdaBoost classifiers can be refined using the newly labeled data, in which one part is labeled by itself and the other part is labeled by another group of TrAdaBoost classifiers. The newly labeled samples are carefully selected under certain conditions, which is verified in tri-training \citep{11}. Experiments on several UCI data sets and the text classification task data set illustrate the effectiveness of Co-Transfer.

Note that in this paper only the decision tree is used as base classifier in Co-Transfer. In addition, other well-known algorithms, such as Naive Bayes, SVM, and Neural Networks etc, can also be employed as base classifier. In future, these methods can be utilized to test the performance of Co-Transfer with different base classifiers. And further, more data sets, especially the data sets from real applications, should be used to extensively evaluate Co-Transfer. Transferability between the source and target domains is also an important issue for Co-Transfer.

\section*{Acknowledgements}
This work was partially supported by the National Natural Science Foundation of China (61866007), the Natural Science Foundation of Guangxi District (2018GXNSFDA138006), Guangxi Key Laboratory of Trusted Software (KX201721), Image and graphic intelligent processing project of Key Laboratory Fund (GIIP2005,GIIP201505).

\bibliography{sample-base1}
\bibliographystyle{iclr2021_conference}

\end{document}